\documentclass[11pt]{article}

\usepackage{fontspec}
\usepackage{amsmath,amssymb}
\usepackage{booktabs}
\usepackage{longtable}
\usepackage{array}
\usepackage{calc}
\usepackage{graphicx}
\usepackage{hyperref}
\usepackage{xcolor}
\usepackage{fancyvrb}
\fvset{fontsize=\small}

\usepackage[margin=1in]{geometry}

\providecommand{\tightlist}{%
  \setlength{\itemsep}{0pt}\setlength{\parskip}{0pt}}

\makeatletter
\def\maxwidth{\ifdim\Gin@nat@width>\linewidth\linewidth\else\Gin@nat@width\fi}
\def\maxheight{\ifdim\Gin@nat@height>\textheight\textheight\else\Gin@nat@height\fi}
\makeatother
\setkeys{Gin}{width=\maxwidth,height=\maxheight,keepaspectratio}

\title{From Scalars to Tensors: Declared Losses Recover Epistemic
  Distinctions That Neutrosophic Scalars Cannot Express}

\author{
  Tony Mason \\[4pt]
  University of British Columbia, Vancouver, Canada \\
  Georgia Institute of Technology, Atlanta, USA \\
  \texttt{fsgeek@cs.ubc.ca; fsgeek@gatech.edu, fsgeek@wamson.com}
}

\date{}

\begin{document}
\maketitle

\begin{abstract}

Leyva-Vázquez and Smarandache (2025) demonstrated that neutrosophic
T/I/F evaluation, where Truth, Indeterminacy, and Falsity are
independent dimensions not constrained to sum to 1.0, which reveals
``hyper-truth'' (T+I+F > 1.0) in 35\% of complex epistemic
cases evaluated by LLMs. We extend their work in two directions. First,
we replicate and extend their experiment across five model families from
five vendors (Anthropic, Meta, DeepSeek, Alibaba, Mistral), finding
hyper-truth in 84\% of unconstrained evaluations, which confirms the
phenomenon is cross-vendor under our prompt protocol. Second, and
more significantly, we identify a limitation of scalar T/I/F that their
framework cannot address: models adopting an ``Absorption'' position
(T=0, I=1, F=0) produce identical scalar outputs for fundamentally
different epistemic situations (paradox, ignorance, contingency),
collapsing the very distinctions neutrosophic logic was designed to
preserve. We demonstrate that extending the evaluation to include
\textbf{declared losses} (structured descriptions of what the model
cannot evaluate and why) substantially recovers these distinctions. Models
producing identical scalars for paradox and ignorance produce nearly
disjoint loss vocabularies (Jaccard similarity < 0.10 on loss
description keywords), with domain-specific, severity-rated loss
declarations that differentiate the nature of their uncertainty. This
suggests that scalar T/I/F is a necessary but insufficient
representation of epistemic state, and that tensor-structured output
(scalars + losses) provides a more faithful model of LLM epistemic
capabilities.

\end{abstract}

\section{Introduction}

The application of Neutrosophic Logic to LLM uncertainty quantification
(\cite{leyva2025neutrosophic}) represents an important departure
from probabilistic frameworks. By treating Truth (T), Indeterminacy (I),
and Falsity (F) as independent dimensions, the neutrosophic approach
allows models to express internal conflicts that probabilistic
constraints (T+I+F=1) suppress. Their finding that 35\% of evaluations
exhibit ``hyper-truth'' (T+I+F > 1.0) under unconstrained
prompting demonstrates that LLMs carry richer epistemic information than
probabilistic output formats can express.

However, their study has three limitations that constrain the
generalizability of their findings:

\begin{enumerate}
\def\labelenumi{\arabic{enumi}.}
\item
  \textbf{Single vendor}: All four models (GPT-4o, GPT-4-turbo,
  GPT-3.5-turbo, GPT-4o-mini) come from OpenAI. The observed hyper-truth
  could be an artifact of OpenAI's training pipeline rather than a
  general property of LLMs.
\item
  \textbf{Single shot}: Each cell (model × phenomenon × strategy) was
  evaluated once, providing no measure of intra-model consistency or
  statistical significance.
\item
  \textbf{Unpublished prompts}: The specific prompts used for each
  strategy were not committed to the repository, preventing independent
  replication.
\end{enumerate}

We address all three limitations. More importantly, we identify a fourth
problem that scalar T/I/F cannot solve even in principle: \textbf{the
Absorption problem}, where some models collapse all uncertain states to
(T=0, I=1, F=0), losing the distinction between types of uncertainty
that neutrosophic logic was designed to preserve.

We then demonstrate that this problem is solved by extending the output
format from scalars to tensors: specifically, by requiring models to
declare structured losses alongside their T/I/F values.

\subsection{Related Work}

\paragraph{Neutrosophic logic foundations.}
The formal basis for independent truth, indeterminacy, and falsity
dimensions comes from neutrosophic logic and set theory
(\cite{smarandache2010neutrosophic,smarandache2005unifying}),
with subsequent critical analysis of expressiveness and
interpretability tradeoffs (\cite{rivieccio2008neutrosophic}).
Our work operates within this tradition but focuses on an empirical
question for modern LLMs: what epistemic distinctions current output
formats preserve versus collapse.

\paragraph{Neutrosophic uncertainty in LLMs.}
The immediate empirical precursor is Leyva-Vázquez and Smarandache's
LLM study (\cite{leyva2025neutrosophic}), which introduced S1--S3
prompting strategies and reported hyper-truth under unconstrained
evaluation. We extend this line with cross-vendor repeated trials and a
new S4 tensor-output protocol designed to expose representational loss
within scalar T/I/F outputs.

\paragraph{Adjacent uncertainty frameworks in LLMs.}
Recent work on abstention and risk control in LLM systems emphasizes
calibration, selective prediction, and refusal policies
(\cite{yadkori2024mitigating,tayebati2025learning,machcha2026knowing}).
This literature asks when a model should abstain. Our question is
orthogonal: when a model does provide a T/I/F assessment, how much
epistemic structure is lost by restricting output to three scalars.

\paragraph{Honesty and self-knowledge.}
Research on honesty, overconfidence mitigation, and reasoning-linked
reliability (\cite{li2024survey,yanlearn,ahtisham2026llm})
is closely related to our ``declared losses'' mechanism. We treat losses
as structured self-reports of epistemic limits and evaluate whether
those reports carry discriminative content beyond scalar uncertainty.

\section{Cross-Vendor Replication}

\subsection{Method}

We designed prompts for all three strategies described by Leyva-Vázquez
and Smarandache (see Appendix A for full prompts):

\begin{itemize}
\tightlist
\item
  \textbf{S1 (Neutrosophic)}: Independent T, I, F on {[}0,1{]},
  explicitly stated as not constrained to sum to 1.0
\item
  \textbf{S2 (Probabilistic)}: T + I + F constrained to 1.0
\item
  \textbf{S3 (Entropy-Derived)}: Binary P(yes)/P(no), indeterminacy
  derived from Shannon entropy~\cite{shannon1948mathematical}:
  H = -(p$\cdot$log$_{2}$p + (1-p)$\cdot$log$_{2}$(1-p)). Note:
  since T=P\_yes and F=P\_no with T+F=1.0 by construction, S3 Sum = 1.0
  + H, meaning S3 produces hyper-truth by mathematical construction
  whenever P\_yes is not exactly 0 or 1. S3 hyper-truth rates are
  therefore not directly comparable to S1 and are not used in our
  analysis
\end{itemize}

We evaluated five models from five vendors via OpenRouter's
OpenAI-compatible API:

\begin{longtable}[]{@{}lll@{}}
\toprule\noalign{}
Model & Provider & Parameters \\
\midrule\noalign{}
\endhead
\bottomrule\noalign{}
\endlastfoot
Claude Sonnet 4.6 & Anthropic & Undisclosed \\
Llama 4 Maverick & Meta & Undisclosed \\
DeepSeek V3 & DeepSeek & 671B MoE \\
Qwen3-235B & Alibaba & 235B MoE \\
Mistral Medium 3.1 & Mistral & Undisclosed \\
\end{longtable}

Each cell was evaluated 5 times (temperature=0.7), yielding 375
evaluations (5 models × 5 phenomena × 3 strategies × 5 reps). Of these,
373 produced valid JSON parses; 2 responses were garbled or truncated (1
DeepSeek, 1 Llama) and are excluded from analysis. All data, code, and
prompts are published in the repository.

\textbf{Important caveat}: Because the original prompts were not
published, we designed our prompts from the strategy descriptions in the
paper. We cannot claim direct replication, which is a parallel
experiment with independently constructed prompts. The S2 control
(below) provides evidence that our prompt framing is not radically
different from the original.

\subsection{Results}

\textbf{S2 constraint validation.} Under probabilistic constraint (S2),
all 125 evaluations across all models and phenomena produced Sum = 1.000
± 0.000, with 0\% hyper-truth. This matches the original study exactly
and provides evidence that our prompt construction is compatible with
theirs: the constraint mechanism behaves identically.

\textbf{S1 hyper-truth is cross-vendor and amplified.} Under
unconstrained neutrosophic evaluation (S1):

\begin{longtable}[]{@{}llll@{}}
\toprule\noalign{}
Model & Hyper\% & Mean Sum & CV(Sum) \\
\midrule\noalign{}
\endhead
\bottomrule\noalign{}
\endlastfoot
Claude Sonnet 4.6 & 100\% & 1.810 & 0.110 \\
DeepSeek V3 & 100\% & 1.475 & 0.161 \\
Qwen3-235B & 80\% & 1.464 & 0.242 \\
Llama 4 Maverick & 76\% & 1.348 & 0.230 \\
Mistral Medium 3.1 & 64\% & 1.312 & 0.281 \\
\end{longtable}

Overall S1 hyper-truth: 84\% (104/124 valid cells; 1 garbled response
excluded), compared to the original study's 35\% (7/20 cells). Because
our prompts were independently constructed, this gap should be read as a
cross-study difference rather than a direct effect-size comparison.
Hyper-truth is not an OpenAI artifact as it emerges across all five
vendor families.

\textbf{Phenomenon-level comparison with original:}

\begin{longtable}[]{@{}llll@{}}
\toprule\noalign{}
Phenomenon & Original Sum & Ours (S1) & Delta \\
\midrule\noalign{}
\endhead
\bottomrule\noalign{}
\endlastfoot
Paradox (Logical) & 1.500 & 1.480 & -0.020 \\
Ignorance (Epistemic) & 1.125 & 1.526 & +0.401 \\
Vagueness (Fuzzy) & 1.125 & 1.382 & +0.257 \\
Contradiction (Ethical) & 1.475 & 1.690 & +0.215 \\
Contingency (Future) & 1.000 & 1.325 & +0.325 \\
\end{longtable}

Ethical contradiction achieves 100\% hyper-truth across all models in
our data, confirming the original study's finding that moral conflict is
the strongest driver of hyper-truth.

\subsection{Three Philosophical Positions on Paradox}

The liar's paradox (``This sentence is false'') reveals three distinct
interpretations of T/I/F. Three of five models (Claude, DeepSeek, Llama)
show zero intra-model variance across all 5 repetitions. Mistral is
near-consistent (4/5 reps identical, 1 outlier). Qwen shows position
instability, adopting different positions across reps.

\textbf{Position 1: Saturation} (Claude, 5/5 reps): T=0.5, I=1.0, F=0.5,
Sum=2.0. The paradox is simultaneously half-true, half-false, and
maximally indeterminate. All three dimensions are active independently.

\textbf{Position 2: Balanced Conflict} (DeepSeek, 5/5 reps): T=0.5,
I=0.5, F=0.5, Sum=1.5. Equal truth and falsity with moderate
indeterminacy. The paradox creates contradiction but not maximal
uncertainty.

\textbf{Position 3: Absorption} (Llama 5/5, Mistral 4/5): T=0.0, I=1.0,
F=0.0, Sum=1.0. Indeterminacy absorbs truth value entirely. The paradox
is treated as having no truth value, effectively the classical logic
response of rejecting the statement as malformed.

Qwen alternates between Positions 1 and 3, suggesting the model has not
converged on a stable interpretation of the paradox.

\begin{figure}
\centering
\includegraphics{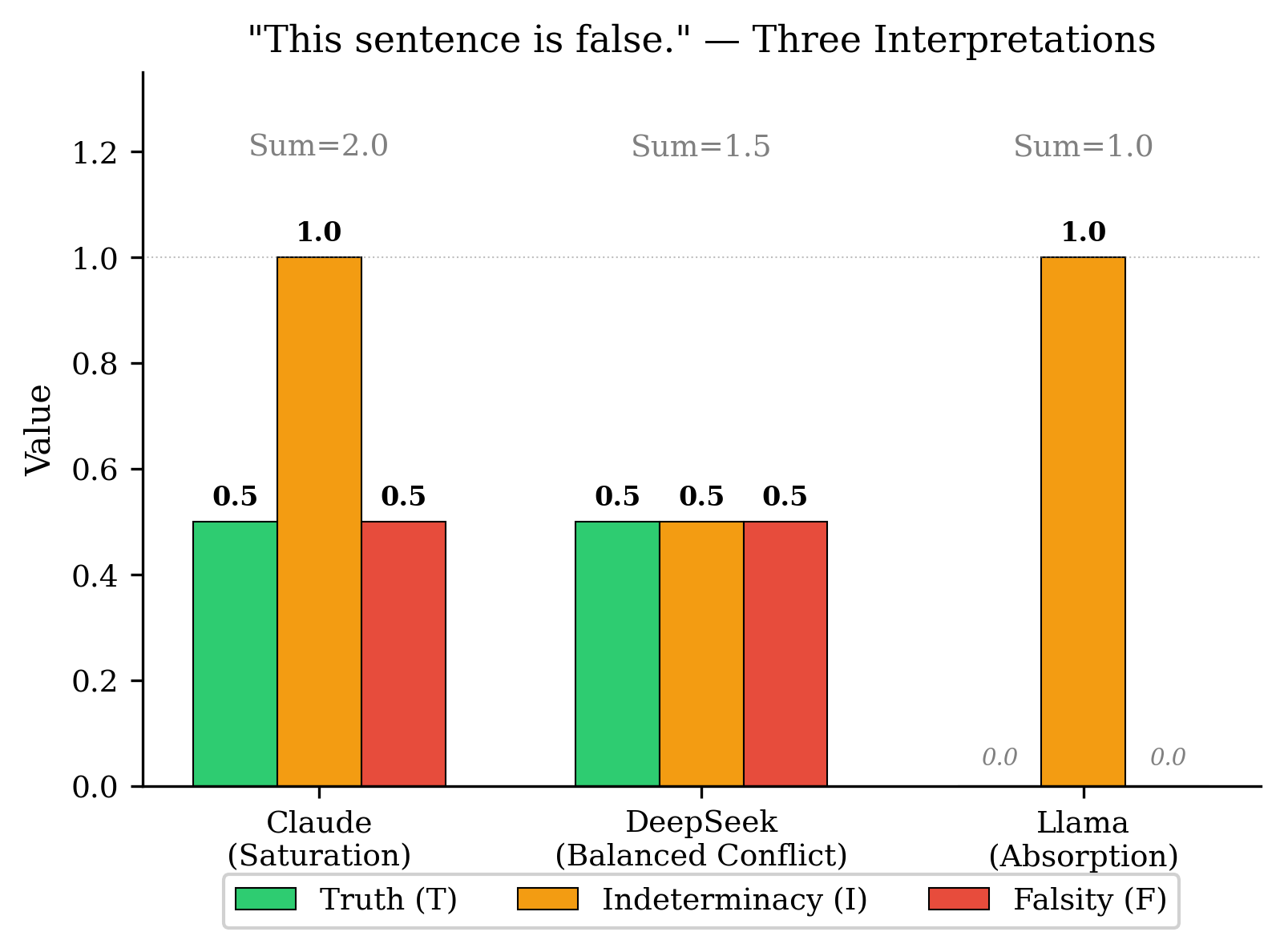}
\caption{Figure 1: Three philosophical positions on the liar's paradox}
\end{figure}

\emph{Figure 1: T/I/F values for ``This sentence is false'' across three
models exhibiting distinct interpretations. Absorption (Llama) collapses
T and F to zero, leaving only I --- the same scalar output as ignorance
and contingency.}

These positions are invisible to any metric based on Sum alone. The
original study, which reported only sum-based analysis, could not have
detected them.

Critically, the Absorption position (Position 3) creates a problem for
neutrosophic logic itself: models predominantly adopting this position
produce near-identical T/I/F scalars for paradox, ignorance, and
contingency, which are three fundamentally different types of
uncertainty. The scalar representation collapses exactly the
distinctions neutrosophic logic was designed to preserve.

\section{The Absorption Problem}

\subsection{Definition}

We define the \textbf{Absorption problem} as the case where a model maps
multiple distinct epistemic states to the same scalar T/I/F vector,
typically (T=0, I$\approx$1, F=0). Under Absorption, indeterminacy absorbs the
truth and falsity dimensions, leaving the scalar output unable to
distinguish between types of uncertainty.

\subsection{Evidence}

Mistral Medium 3.1 predominantly exhibits Absorption across multiple
phenomena in S1 (modal response across 5 reps shown):

\begin{longtable}[]{@{}llllll@{}}
\toprule\noalign{}
Phenomenon & T & I & F & Sum & Consistency \\
\midrule\noalign{}
\endhead
\bottomrule\noalign{}
\endlastfoot
Paradox (Logical) & 0.00 & 1.00 & 0.00 & 1.00 & 4/5 reps \\
Ignorance (Epistemic) & 0.00 & 1.00 & 0.00 & 1.00 & 3/5 reps \\
Contingency (Future) & 0.32 & 0.71 & 0.11 & 1.14 & mean values \\
\end{longtable}

The original study's flagship model (GPT-4o) shows the same pattern:
(T=0, I=1, F=0) for both paradox and ignorance. Neither Leyva-Vázquez
nor Smarandache discuss this.

Absorption is not a model deficiency: it is a representational
limitation. The model may have richer internal distinctions that the
scalar output format cannot express. Testing this hypothesis motivates
the tensor extension.

\section{Strategy 4: Tensor Extension with Declared Losses}

\subsection{Method}

We designed a fourth evaluation strategy (S4) that extends S1 by
requiring the model to declare \textbf{structured losses} alongside its
T/I/F values. Each loss is a triple:

\begin{itemize}
\tightlist
\item
  \textbf{what}: What the model cannot evaluate (brief description)
\item
  \textbf{why}: Why this limits the assessment
\item
  \textbf{severity}: Impact on the evaluation {[}0.0 to 1.0{]}
\end{itemize}

The prompt explicitly requires at least one declared loss and states
that ``honesty about limits is required.'' Full prompt text in Appendix
A.

We ran S4 across all 5 models, 5 phenomena, and 5 repetitions (125
evaluations) with max\_tokens=500. This proved insufficient for
Mistral's verbose loss declarations, causing 18 parse failures from
truncated JSON. Mistral was re-run at max\_tokens=1500, producing 25
complete evaluations with 0 parse failures. The Mistral results reported
below use the 1500-token rerun data exclusively.

\subsection{The Key Test: Do Losses Differentiate What Scalars Cannot?}

For each model, we computed two metrics comparing paradox vs.~ignorance:

\begin{enumerate}
\def\labelenumi{\arabic{enumi}.}
\tightlist
\item
  \textbf{Scalar Manhattan distance}: \textbar T\_p - T\_i\textbar{} +
  \textbar I\_p - I\_i\textbar{} + \textbar F\_p - F\_i\textbar{} where
  p = paradox, i = ignorance
\item
  \textbf{Loss vocabulary Jaccard similarity}: Word-level token overlap
  between the ``what'' fields of declared losses for the two phenomena
  (the concise loss description, not the explanatory ``why'' field)
\end{enumerate}

\begin{longtable}[]{@{}
  >{\raggedright\arraybackslash}p{(\columnwidth - 6\tabcolsep) * \real{0.3182}}
  >{\centering\arraybackslash}p{(\columnwidth - 6\tabcolsep) * \real{0.1364}}
  >{\centering\arraybackslash}p{(\columnwidth - 6\tabcolsep) * \real{0.1364}}
  >{\raggedright\arraybackslash}p{(\columnwidth - 6\tabcolsep) * \real{0.4091}}@{}}
\toprule\noalign{}
\begin{minipage}[b]{\linewidth}\raggedright
Model
\end{minipage} & \begin{minipage}[b]{\linewidth}\centering
Scalar Distance
\end{minipage} & \begin{minipage}[b]{\linewidth}\centering
Loss Jaccard
\end{minipage} & \begin{minipage}[b]{\linewidth}\raggedright
Finding
\end{minipage} \\
\midrule\noalign{}
\endhead
\bottomrule\noalign{}
\endlastfoot
Claude Sonnet 4.6 & 0.034 & 0.097 & Scalars barely distinguish; losses
fully distinguish \\
DeepSeek V3 & 0.540 & 0.083 & Both channels distinguish \\
Llama 4 Maverick & 0.040 & 0.056 & Scalars nearly identical; losses
disjoint \\
Mistral Medium 3.1 & \textbf{0.000} & \textbf{0.066} & \textbf{Scalars
identical; losses completely different} \\
Qwen3-235B & 0.340 & 0.070 & Both channels distinguish \\
\end{longtable}

\textbf{Every model produces nearly disjoint loss vocabularies} for
paradox vs.~ignorance (Jaccard < 0.10), regardless of whether
the scalars distinguish the phenomena. The maximum vocabulary overlap is
9.7\% (Claude). The minimum is 5.6\% (Llama).

\begin{figure}
\centering
\includegraphics{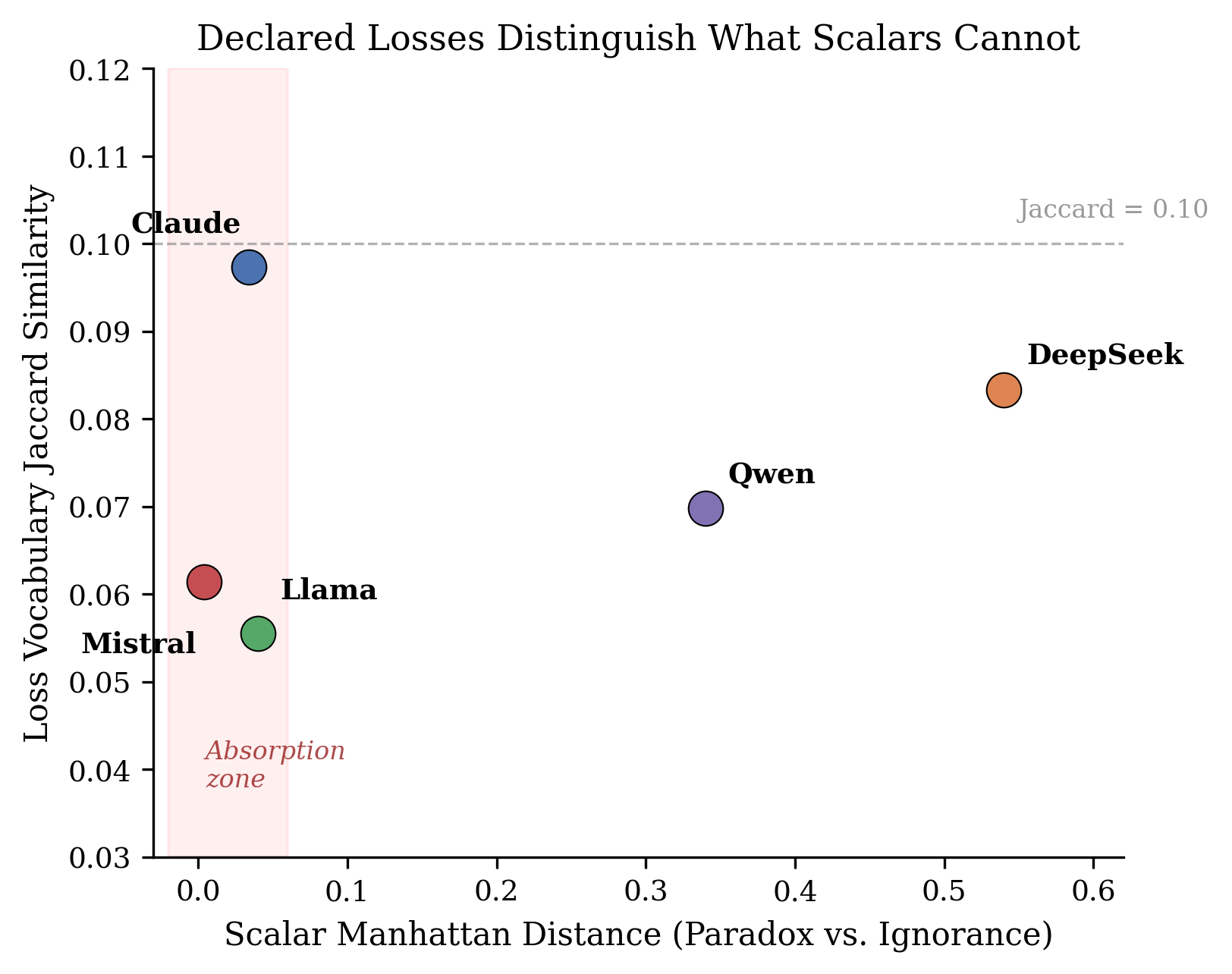}
\caption{Figure 2: Scalar distance vs.~loss Jaccard similarity}
\end{figure}

\emph{Figure 2: Scalar Manhattan distance (x-axis) vs.~loss vocabulary
Jaccard similarity (y-axis) for paradox vs.~ignorance across five
models. Models in the Absorption zone (red shading) produce near-zero
scalar differences yet completely different loss declarations. All
models fall below Jaccard = 0.10.}

\subsection{Mistral: The Critical Case}

Mistral is the strongest test because it predominantly exhibits
Absorption by producing (T=0, I=1, F=0) as its modal response for both
paradox and ignorance in S1 and S4:

\textbf{S4 Paradox} (T=0.0, I=1.0, F=0.0): ``Self-referential paradox
resolution'' (severity 1.0), ``Formal system dependency'' (0.8),
``Contextual grounding of `this sentence'\,'' (1.0)

\textbf{S4 Ignorance} (T=0.0, I=1.0, F=0.0): ``Empirical unknowability
of the exact number of stars'' (severity 1.0), ``Definition of
`universe' in cosmological context'' (0.9), ``Mathematical ambiguity of
`even' for infinite quantities'' (0.8)

The scalars are identical. The losses are domain-specific, accurate, and
completely different. The model \textbf{has} the internal distinction:
the scalar output format cannot express it.

\subsection{Pairwise Loss Differentiation (Mistral)}

The full pairwise Jaccard matrix for Mistral across all five phenomena
(from the 1500-token rerun, 25 evaluations, 0 parse failures):

\begin{longtable}[]{@{}llllll@{}}
\toprule\noalign{}
& Par & Ign & Vag & Con & Fut \\
\midrule\noalign{}
\endhead
\bottomrule\noalign{}
\endlastfoot
\textbf{Paradox} & 1.000 & 0.061 & 0.089 & 0.126 & 0.074 \\
\textbf{Ignorance} & & 1.000 & 0.089 & 0.116 & 0.084 \\
\textbf{Vagueness} & & & 1.000 & 0.143 & 0.091 \\
\textbf{Contradiction} & & & & 1.000 & 0.099 \\
\textbf{Contingency} & & & & & 1.000 \\
\end{longtable}

Every off-diagonal cell is below 0.15. Five phenomena, five nearly
disjoint loss vocabularies from a model that produces nearly identical
scalars for three of them.

\begin{figure}
\centering
\includegraphics{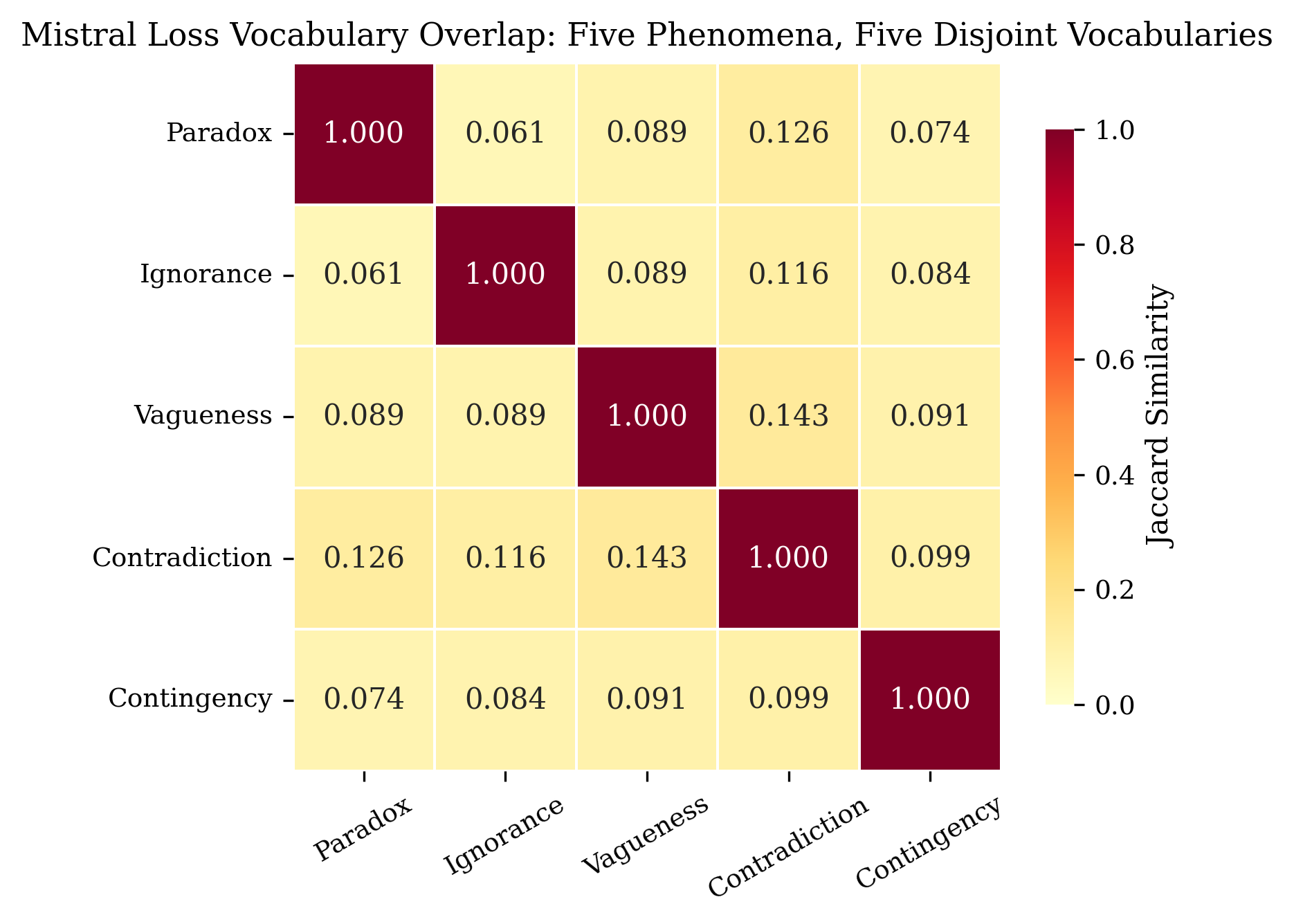}
\caption{Figure 3: Mistral pairwise Jaccard heatmap}
\end{figure}

\emph{Figure 3: Pairwise Jaccard similarity of Mistral's loss
vocabularies across all five phenomena. The hot diagonal
(self-similarity = 1.0) against uniformly cold off-diagonal cells (all
< 0.15) shows five nearly disjoint loss vocabularies from a
model that produces near-identical scalars.}

\subsection{Severity Profiles Add Further Discrimination}

Mean loss severity varies by phenomenon, even when scalars do not:

\begin{longtable}[]{@{}lcc@{}}
\toprule\noalign{}
Phenomenon & Mean Severity & Avg Losses/Rep \\
\midrule\noalign{}
\endhead
\bottomrule\noalign{}
\endlastfoot
Paradox & 0.795 & 4.2 \\
Contingency & 0.779 & 5.2 \\
Ignorance & 0.760 & 5.0 \\
Contradiction & 0.748 & 5.0 \\
Vagueness & 0.510 & 5.2 \\
\end{longtable}

Vagueness has markedly lower severity (0.510 vs 0.748--0.795),
reflecting that fuzzy boundary uncertainty is less severe than paradox
or ignorance. This is an appropriate distinction that the scalar
representation completely misses.

\subsection{Validation: Instruction-Following vs.\ Epistemic Assessment}
\label{sec:validation}

An adversarial reading of S4 might dismiss the results as ``just
instruction-following'': given a prompt that requests T/I/F values and
loss declarations, models dutifully produce both, and the apparent
structure is an artifact of compliance rather than evidence of internal
epistemic assessment.  We test this interpretation with four analyses
of the existing S4 data and two new experiments (72 additional API
calls).

\subsubsection{Test 1: Severity--Indeterminacy Correlation}

Across all 125 S4 evaluations, the maximum declared loss severity per
response correlates strongly with the corresponding I value: Pearson
$r = 0.78$ ($p < 10^{-26}$), Spearman $\rho = 0.83$
($p < 10^{-32}$).  The relationship holds within each of the five
models individually (all $p < 0.001$).  The S4 prompt requests T, I, F
values and loss declarations as independent output fields---it does not
instruct models to make severity correlate with indeterminacy.  Yet all
five models produce this internal coherence unprompted.

The mean--max distinction matters.  Mean severity shows a moderate
correlation ($r = 0.43$), but this is explained by stimulus identity:
harder stimuli produce both higher mean severity and higher~I.  After
residualizing by stimulus, the mean severity correlation disappears
($r = 0.02$, $p = 0.81$).  We report this honestly: the mean
severity relationship is a stimulus-level confound, not a within-cell
regularity.  Max severity, by contrast, survives stimulus
residualization ($r = 0.54$, $p < 10^{-10}$) and double
residualization removing both stimulus \emph{and} model effects
($r = 0.29$, $p < 0.001$).  The severity--indeterminacy coupling is
a within-cell structural relationship that cannot be attributed to
stimulus difficulty or model-level intercept differences.

\subsubsection{Test 2: Cross-Model Theme Convergence}

If loss declarations were templated compliance, different models would
produce either identical boilerplate or uncorrelated noise.  Instead, we
observe structured convergence: 8 epistemic themes achieved universal
agreement across all 5 models (mean pairwise Jaccard similarity 0.450;
permutation test $p < 0.0001$ for non-random convergence).

Theme content is stimulus-specific.  The liar paradox elicits
self-reference and bivalence-limits.  Rain elicits temporal uncertainty
and empirical-access limitations.  Stars elicits scope and
measurement-precision concerns.  Across all models and repetitions, the
S4 data contain 324 unique loss descriptions out of 417 total
declarations, with pairwise Jaccard similarity of exactly 0.000 between
all 10 stimulus pairs: not a single loss description appears in more
than one stimulus.

Models differ in depth but not direction.  Claude and Mistral produce
expansive declarations (4--6 losses per response); Llama and DeepSeek
produce efficient ones (2--3 losses).  All converge on the same themes
per stimulus.  This is stimulus-specific content generation with
cross-model structural agreement, not templated output.

\subsubsection{Test 3: Tautology Control}
\label{sec:tautology}

If models merely comply with format, they should produce similar I
values and severity scores for tautologies as for genuinely uncertain
stimuli---the prompt makes the same requests either way.  We tested
three tautologies:

\begin{enumerate}
\def\labelenumi{\arabic{enumi}.}
\tightlist
\item
  ``2+2=4'' (mathematical)
\item
  ``All bachelors are unmarried'' (definitional)
\item
  ``It is raining or it is not raining'' (logical)
\end{enumerate}

\noindent
across three models (Claude Sonnet 4.6, DeepSeek V3, Llama~4 Maverick),
3 repetitions each (27 API calls total).  Table~\ref{tab:tautology}
shows the comparison.

\begin{table}[t]
\centering
\caption{S4 responses for tautologies vs.\ original stimuli. Values are
means across all models and repetitions.}
\label{tab:tautology}
\begin{tabular}{lccc}
\toprule
& \textbf{I} & \textbf{Max Severity} & \textbf{Losses/Rep} \\
\midrule
Original stimuli ($n=125$)  & 0.739 & 0.740 & 3.0 \\
Tautologies ($n=27$)        & 0.031 & 0.101 & 2.0 \\
\midrule
\textbf{Ratio (taut/orig)}  & \textbf{0.04$\times$} & \textbf{0.14$\times$} & 0.67$\times$ \\
\bottomrule
\end{tabular}
\end{table}

Indeterminacy drops by 25$\times$ and maximum severity by 7$\times$.
Loss count, however, drops only modestly (3.0 to 2.0): models comply
with the format requirement of declaring at least one loss but
\emph{modulate the content}.  Tautology losses are minimal-severity
hedges (``language ambiguity,'' ``formal system assumptions''), not the
substantive epistemic limitations declared for the original stimuli.

Two model-specific behaviors are instructive.  Claude refused $T = 1.0$
for any tautology, retaining a sliver of I (e.g., $I = 0.05$
for ``2+2=4'').  DeepSeek collapsed to exactly one loss at minimum
severity.  The logical tautology (``raining or not raining'') produced
slightly higher I values than the mathematical one---appropriate, since
excluded middle has genuine philosophical challenges that ``2+2=4'' does
not.\footnote{Tautology data: \texttt{data/tautology\_s4\_results.json}.}

\subsubsection{Test 4: Prompt Ablation}
\label{sec:ablation}

Tests 1--3 show that S4 responses are internally coherent and
content-modulated.  But they leave open the question of whether the
neutrosophic framing \emph{elicits} the epistemic assessment or merely
\emph{channels} it.  To test this, we stripped all neutrosophic
apparatus and ran a minimal prompt (S5):

\begin{quote}
\emph{``Identify any limitations, uncertainties, or difficulties in
determining whether this statement is true or false.''}
\end{quote}

\noindent
No T/I/F, no severity scores, no ``MUST declare losses.''  Same 5
stimuli, same 3 models, 3 repetitions each (45 API calls).
Table~\ref{tab:ablation} shows theme overlap between the S5 free-form
responses and the 10 reference themes identified in S4.

\begin{table}[t]
\centering
\caption{S5 (ablated prompt) theme overlap with S4 reference themes.
$\checkmark$ = theme present in all 3 repetitions;
\textsuperscript{$\circ$} = present in 2/3.
All 10 S4 reference themes appear in every model.}
\label{tab:ablation}
\begin{tabular}{lccc}
\toprule
\textbf{S4 Reference Theme} & \textbf{Claude} & \textbf{DeepSeek} & \textbf{Llama} \\
\midrule
Self-reference           & $\checkmark$ & $\checkmark$ & $\checkmark$ \\
Bivalence limits         & $\checkmark$ & $\checkmark$ & $\checkmark$ \\
Temporal uncertainty     & $\checkmark$ & $\checkmark$ & $\checkmark$ \\
Empirical access         & $\checkmark$ & $\checkmark$ & $\checkmark$ \\
Scope/cardinality        & $\checkmark$ & $\checkmark$ & $\checkmark$ \\
Measurement precision    & $\checkmark$ & $\checkmark$ & $\checkmark$ \\
Cultural norms           & $\checkmark$ & $\checkmark$ & $\checkmark$ \\
Contextual definitions   & $\checkmark$ & $\checkmark$ & $\checkmark$ \\
Moral pluralism          & $\checkmark$ & $\checkmark$ & $\checkmark$ \\
Consequentialist tension & $\checkmark$ & $\checkmark$ & $\checkmark$ \\
\bottomrule
\end{tabular}
\end{table}

The result is 100\% theme overlap: all 10 S4 reference themes appeared
in every model, every repetition, under the ablated prompt.  The
neutrosophic framing is not required to elicit the epistemic content.

S5 also produced 6 additional themes absent from S4:
language-ambiguity (42\% of responses), pragmatic-truth (22\%),
observer-dependence, incompleteness, logical-framework-choice, and
temporal-change.  Crucially, these are not evaluation-relevant losses
but \emph{task-preconditions}---observations about the difficulty of
evaluating rather than factors that affect the evaluation scores.
Language-ambiguity in ``it will rain tomorrow'' is real, but once one
has fixed the meanings of ``rain'' and ``tomorrow,'' the T/I/F
assessment proceeds without it.  The 6 extra themes describe conditions
that must be resolved \emph{before} evaluation begins, not uncertainty
that persists \emph{within} it.  Their absence from S4 output is
evidence that the formalism is working as intended: constraining the
observation space to what is relevant to the T/I/F
assessment.\footnote{Ablation data: \texttt{data/s5\_ablation\_results.json}.}

\subsubsection{Scope Confirmation}

The ablation reveals that S4 is a \emph{channel}, not a
\emph{source}.  The epistemic assessment pre-exists the prompt format;
S4 quantifies and standardizes it.  The 6 additional themes S5
surfaces are epistemically interesting but fall outside S4's scope
\emph{by design}: they are meta-level observations about the evaluation
task itself (``language is ambiguous,'' ``logical frameworks differ''),
not products of performing the evaluation.  This is what formal models
do---they exclude the irrelevant.  The S5 ablation confirms that S4
captures 100\% of evaluation-relevant themes (the 10 shared themes)
while correctly excluding preconditions that do not bear on T/I/F
scores.

S4's value is quantification and cross-model comparability: it converts
qualitative assessment into comparable tensors across models and stimuli.
The ablation confirms the scope of that conversion rather than exposing
a gap in it.

\subsubsection{Summary of the Validation Battery}

S4 also compresses I-value variance by 3--38$\times$ relative to
S1--S3 (38$\times$ for epistemic ignorance: S4 variance 0.0029 vs.\
S1--S3 variance 0.1112).  Requiring loss declarations anchors the
scalar assessment rather than decorating it.  Table~\ref{tab:battery}
summarizes the four tests.

\begin{table}[t]
\centering
\caption{Validation battery summary. Each test addresses a different
facet of the instruction-following objection.}
\label{tab:battery}
\begin{tabular}{lp{5.2cm}l}
\toprule
\textbf{Test} & \textbf{Finding} & \textbf{Against} \\
\midrule
Severity--I correlation & $r = 0.29$ after double residualization ($p < 0.001$)
  & Uncoupled compliance \\
Theme convergence & 8 universal themes, Jaccard = 0.000 cross-stimulus
  & Templated output \\
Tautology control & I drops 25$\times$, severity 7$\times$
  & Format-driven inflation \\
Prompt ablation & 100\% theme overlap; 6 extra precondition themes correctly outside S4 scope
  & Framing-dependent elicitation \\
\bottomrule
\end{tabular}
\end{table}

A compliant-but-uncoupled response to the S4 prompt would produce loss
severity uncorrelated with I (the prompt does not request coherence),
generic loss descriptions reused across stimuli (the prompt does not
prohibit this), unmodulated output for tautologies (the prompt makes
identical requests), and content dependent on neutrosophic framing (a
different prompt would elicit different themes).  The data show none of
these.  S4 surfaces pre-existing epistemic assessment; the neutrosophic
tensor structures it for comparison while correctly scoping to
evaluation-relevant factors.

\section{Discussion}

\subsection{Scalar T/I/F Is Necessary But Insufficient}

The original Smarandache and Leyva-Vázquez thesis is correct:
independent T/I/F dimensions reveal hyper-truth that probabilistic
constraints suppress. Our cross-vendor parallel experiment extends this
claim across five vendors and reports higher S1 hyper-truth frequency
under our independently constructed prompts.

However, scalar T/I/F has a ceiling. The Absorption problem shows that
some models cannot express the distinction between different types of
uncertainty using three numbers alone. This is not a model deficiency,
it is a representational limitation. The model carries the distinction
internally but lacks the output dimensions to express it.

\subsection{Declared Losses Are the Missing Dimension}

Adding structured loss declarations to the evaluation output recovers
the collapsed distinctions. The loss channel carries semantic
information that the scalar channel drops: \emph{why} the model is
uncertain, \emph{what} it cannot evaluate, and \emph{how severely} this
limits its assessment.

This suggests a hierarchy of epistemic output fidelity:

\begin{enumerate}
\def\labelenumi{\arabic{enumi}.}
\tightlist
\item
  \textbf{Probabilistic} (T+I+F=1): Collapses all uncertainty to a
  single dimension. Cannot express conflict. Cannot express hyper-truth.
\item
  \textbf{Scalar Neutrosophic} (independent T,I,F): Expresses conflict
  and hyper-truth. Cannot distinguish types of uncertainty for
  Absorption models.
\item
  \textbf{Tensor Neutrosophic} (T,I,F + declared losses): Expresses
  conflict, hyper-truth, and type-specific uncertainty. Recovers
  distinctions that scalars collapse.
\end{enumerate}

Each level subsumes the previous. The tensor level is the minimum needed
for faithful representation of LLM epistemic state.

\subsection{Implications for Evaluation Architecture}

The finding that models can produce accurate, domain-specific loss
declarations has practical implications for AI evaluation systems. An
attestation system that records only scalar T/I/F values will lose
information that loss declarations preserve. Systems designed for
routing evaluations to appropriate evaluators by selecting which model
or method to use for a given claim, should route on losses rather than
capabilities, because losses carry richer information about what the
evaluator cannot do and why.

\subsection{Limitations}

\begin{enumerate}
\def\labelenumi{\arabic{enumi}.}
\item
  \textbf{Prompt sensitivity}: We do not have the original study's
  prompts, so the S1--S3 comparison is across independently constructed
  prompts. The S2 agreement provides evidence of compatibility but not
  identity.
\item
  \textbf{Sample size}: 5 repetitions per cell provides stability
  measures but is limited for confirmatory hypothesis testing under
  strict independence assumptions. The statistical tests in
  Section~\ref{sec:validation} should be read as exploratory evidence.
  The
  zero-variance findings (e.g., Claude's paradox position) are robust;
  the variable findings (e.g., Qwen's paradox instability) would benefit
  from 30+ repetitions.
\item
  \textbf{Loss vocabulary metric}: Jaccard similarity on word-level
  tokens is a crude measure of semantic differentiation. More
  sophisticated NLP measures (sentence embeddings, BERTScore) might
  reveal structure that word-level overlap misses or might show that
  some apparently disjoint vocabularies express similar concepts.
\item
  \textbf{Self-report vs.~internal state}: The declared losses are the
  model's self-report of its limitations, not a direct observation of
  internal state. Models might produce plausible-sounding but generic
  loss declarations without genuine internal distinction. The
  domain-specificity and consistency of the losses across repetitions
  argues against this, but independent validation (e.g., via logprob
  analysis) would strengthen the finding.
\item
  \textbf{Five phenomena, five models}: The stimuli set is small
  (inherited from the original study). A broader range of epistemic
  phenomena would test the generality of both the Absorption problem and
  the tensor solution.
\end{enumerate}

\section{Conclusion}

Neutrosophic scalar evaluation (T/I/F) is a genuine advance over
probabilistic uncertainty quantification for LLMs. Our cross-vendor
replication confirms that hyper-truth is a general property of current
LLMs, not an OpenAI-specific artifact, under the prompt protocol used in
this study.

However, scalar T/I/F hits a representational ceiling: the Absorption
problem causes some models to produce identical outputs for
fundamentally different epistemic states. Extending the output to
include structured loss declarations by creating a tensor
representation, recovers much of these collapsed distinctions in our
experiments.

The model has the distinction. The scalar cannot express it. The tensor
can.

\section*{Acknowledgments and AI Use Disclosure}

The author used Claude Opus 4.6 as a research assistant for drafting,
code-generation support, and editorial iteration. Final responsibility
for study design, claims, analysis, and manuscript content rests with
the human author.

\bibliographystyle{plain}
\bibliography{references}

\appendix
\section{Appendix A: Prompt Strategies}

\subsection{S1 (Neutrosophic)}

\textbf{System}: You are an expert in Neutrosophic Logic. You evaluate
statements using three INDEPENDENT dimensions: Truth (T), Indeterminacy
(I), and Falsity (F), each on {[}0.0, 1.0{]}. These dimensions are NOT
constrained to sum to 1.0. A statement can be simultaneously partially
true AND partially false AND partially indeterminate. Respond with ONLY
a JSON object, no other text.

\textbf{User}: Evaluate this statement on three independent dimensions:

Statement: ``\{statement\}''

\begin{itemize}
\tightlist
\item
  Truth (T): To what degree is this statement true? {[}0.0 to 1.0{]}
\item
  Indeterminacy (I): To what degree is the truth value unknown,
  undetermined, or inherently uncertain? {[}0.0 to 1.0{]}
\item
  Falsity (F): To what degree is this statement false? {[}0.0 to 1.0{]}
\end{itemize}

T, I, and F are independent. They need NOT sum to 1.0.

Respond with ONLY: \{``T'': , ``I'': , ``F'': \}

\subsection{S2 (Probabilistic)}

\textbf{System}: You are a probabilistic classifier. You assign
probabilities to three mutually exclusive categories that MUST sum to
exactly 1.0. Respond with ONLY a JSON object, no other text.

\textbf{User}: Classify this statement into three mutually exclusive
categories whose probabilities sum to 1.0:

Statement: ``\{statement\}''

\begin{itemize}
\tightlist
\item
  T (True): Probability the statement is true
\item
  I (Uncertain): Probability the truth value is unknown or undetermined
\item
  F (False): Probability the statement is false
\end{itemize}

CONSTRAINT: T + I + F must equal 1.0

Respond with ONLY: \{``T'': , ``I'': , ``F'': \}

\subsection{S3 (Entropy-Derived)}

\textbf{System}: You are a binary truth estimator. You estimate the
probability that a statement is true (YES) versus false (NO). The two
probabilities must sum to 1.0. Respond with ONLY a JSON object, no other
text.

\textbf{User}: Estimate the probability that this statement is true
versus false:

Statement: ``\{statement\}''

\begin{itemize}
\tightlist
\item
  P\_yes: Probability the statement is true {[}0.0 to 1.0{]}
\item
  P\_no: Probability the statement is false {[}0.0 to 1.0{]}
\end{itemize}

CONSTRAINT: P\_yes + P\_no must equal 1.0

Respond with ONLY: \{``P\_yes'': , ``P\_no'': \}

\emph{Indeterminacy derived from Shannon entropy:} I = -(p$\cdot$log$_{2}$(p) +
(1-p)$\cdot$log$_{2}$(1-p)) where p = P\_yes

\subsection{S4 (Tensor - Declared Losses)}

\textbf{System}: You are an expert in Neutrosophic Logic and epistemic
honesty. You evaluate statements using three INDEPENDENT dimensions:
Truth (T), Indeterminacy (I), and Falsity (F), each on {[}0.0, 1.0{]}.
These dimensions are NOT constrained to sum to 1.0. Crucially, you must
also declare your LOSSES: what you cannot evaluate, what limits your
assessment, and why your indeterminacy value is what it is. Respond with
ONLY a JSON object, no other text.

\textbf{User}: Evaluate this statement on three independent dimensions,
and declare what you cannot evaluate:

Statement: ``\{statement\}''

\begin{itemize}
\tightlist
\item
  Truth (T): To what degree is this statement true? {[}0.0 to 1.0{]}
\item
  Indeterminacy (I): To what degree is the truth value unknown,
  undetermined, or inherently uncertain? {[}0.0 to 1.0{]}
\item
  Falsity (F): To what degree is this statement false? {[}0.0 to 1.0{]}
\item
  losses: A list of objects, each with:

  \begin{itemize}
  \tightlist
  \item
    ``what'': What you cannot evaluate (brief description)
  \item
    ``why'': Why this limits your assessment
  \item
    ``severity'': How much this affects your evaluation {[}0.0 to 1.0{]}
  \end{itemize}
\end{itemize}

T, I, and F are independent. They need NOT sum to 1.0. You MUST declare
at least one loss. Honesty about limits is required.

Respond with ONLY: \{``T'': , ``I'': , ``F'': , ``losses'':
{[}\{``what'': ``'', ``why'': ``'', ``severity'': \}, \ldots{]}\}

\section{Appendix B: Test Stimuli}

The five test stimuli are from Leyva-Vázquez and Smarandache (2025):

\begin{enumerate}
\def\labelenumi{\arabic{enumi}.}
\tightlist
\item
  \textbf{Paradox (Logical)}: ``This sentence is false.''
\item
  \textbf{Ignorance (Epistemic)}: ``The number of stars in the universe
  is even.''
\item
  \textbf{Vagueness (Fuzzy)}: ``John is 1.75 meters tall, therefore John
  is tall.''
\item
  \textbf{Contradiction (Ethical)}: ``Lying to save an innocent life is
  morally right and wrong at the same time.''
\item
  \textbf{Contingency (Future)}: ``It will rain in New York tomorrow.''
\end{enumerate}

\section{Appendix C: Data Availability}

All code, prompts, and data are available at:
\url{https://github.com/fsgeek/neutrosophic-llm-logic}

\begin{itemize}
\tightlist
\item
  \texttt{src/prompts.py}: All four prompt strategies
\item
  \texttt{src/experiment.py}: Experiment runner with strategy selection
\item
  \texttt{data/cross\_vendor\_results.csv}: S1--S3 production data (375
  evaluations)
\item
  \texttt{data/s4\_tensor\_results.csv}: S4 tensor data (125
  evaluations)
\item
  \texttt{data/s4\_mistral\_rerun.csv}: Mistral S4 rerun at 1500
  max\_tokens (25 evaluations)
\item
  \texttt{data/tautology\_s4\_results.json}: Tautology control data (27
  evaluations, Section~\ref{sec:tautology})
\item
  \texttt{data/s5\_ablation\_results.json}: Prompt ablation data (45
  evaluations, Section~\ref{sec:ablation})
\item
  \texttt{data/openai\_neutrosophic\_results.csv}: Original Smarandache
  data
\end{itemize}

\end{document}